\documentclass[letterpaper, 10 pt, conference]{ieeeconf}  % Comment this line out if you need a4paper

\IEEEoverridecommandlockouts                              % This command is only needed if 
\usepackage{graphicx}
\usepackage{amsmath,amssymb,amsfonts}
\usepackage{algorithmic}
\usepackage{textcomp}
\usepackage{multirow}
\usepackage{arydshln}
\usepackage{xcolor}
\def\BibTeX{{\rm B\kern-.05em{\sc i\kern-.025em b}\kern-.08em
    T\kern-.1667em\lower.7ex\hbox{E}\kern-.125emX}}

\title{\LARGE \bf
Driver-Net: Multi-Camera Fusion for Assessing \\ Driver Take-Over Readiness in Automated Vehicles$^*$}

\author{Mahdi Rezaei$^{1, {\dagger}}$ and Mohsen Azarmi$^{1}$% <-this % stops a space
\thanks{$^{{\dagger}}$\, Corresponding author: Mahdi Rezaei {\tt (m.rezaei@leeds.ac.uk)}}
\thanks{$^{1}$M. Rezaei and M. Azarmi are with the Institute for Transport Studies, Computer Vision and Machine Learning Group, University of Leeds, LS2 9JT, United Kingdom {\tt (m.rezaei@leeds.ac.uk)} and {\tt (tsmaz@leeds.ac.uk)}.}%
\thanks{* This work was funded and supported by the European Union’s Research and Innovation Programme and European Commission for the Hi-Drive Project.}
}

\begin{document}

\maketitle
\thispagestyle{empty}
\pagestyle{empty}

%%%%%%%%%%%%%%%%%%%%%%%%%%%%%%%%%%%%%%%%%%%%%%%%%%%%%%%%%%%%%%%%%%%%%%%%%%%%%%%%
\begin{abstract}

Ensuring safe transition of control in automated vehicles requires an accurate and timely assessment of driver readiness. This paper introduces Driver-Net, a novel deep learning framework that fuses multi-camera inputs to estimate driver take-over readiness. Unlike conventional vision-based driver monitoring systems that focus on head pose or eye gaze, Driver-Net captures synchronised visual cues from the driver’s head, hands, and body posture through a triple-camera setup. The model integrates spatio-temporal data using a dual-path architecture, comprising a Context Block and a Feature Block, followed by a cross-modal fusion strategy to enhance prediction accuracy. Evaluated on a diverse dataset collected from the University of Leeds Driving Simulator, the proposed method achieves an accuracy of up to 95.8\% in driver readiness classification. This performance significantly enhances existing approaches and highlights the importance of multimodal and multi-view fusion. As a real-time, non-intrusive solution, Driver-Net contributes meaningfully to the development of safer and more reliable automated vehicles and aligns with new regulatory mandates and upcoming safety standards.
\end{abstract}

%%%%%%%%%%%%%%%%%%%%%%%%%%%%%%%%%%%%%%%%%%%%%%%%%%%%%%%%%%%%%%%%%%%%%%%%%%%%%%%%
\section{INTRODUCTION}

Driver monitoring systems (DMS) are considered a critical part of new vehicles due to their role in enhancing the broader adoption of Autonomous Vehicles (AV) \cite{DMS}. Such complex systems aim to monitor the driver’s level of readiness by detecting signs of distraction, fatigue, or driver impairments that can affect road safety. 

The importance of DMS technology is constantly increasing, particularly as the safety regulatory organisations, such as the European New Car Assessment Programme (Euro NCAP), have defined a new set of mandatory tests and benchmarks for the inclusion of advanced driver assistance systems (ADAS), including DMS, to gain high safety ratings. 

Since 2020, Euro NCAP has considered awarding points for vehicles equipped with systems capable of detecting driver inattention, distraction, or drowsiness. According to the European Union's General Safety Regulation (GSR), every new vehicle manufactured from 7 July 2024 must be equipped with Driver Drowsiness and Attention Warning (DDAW) systems. By 7 July 2026, Advanced Driver Distraction Warning (ADDW) systems will also be mandatory \cite{GSR}. The objective is not only to reduce accidents caused by human error, which accounts for over 90\% of road crashes globally \cite{WHO}, but also to prepare the automotive industry for future roads dominated by semi-autonomous and autonomous vehicles. 

The United States regulatory bodies, such as the National Highway Traffic Safety Administration (NHTSA), and Asian safety organisations try to align with such new safety requirements and standards which further emphasises the global importance of DMS technologies \cite{NHTSA}.
\begin{figure}[t!]
   \centering
   \includegraphics[width = 1\linewidth]{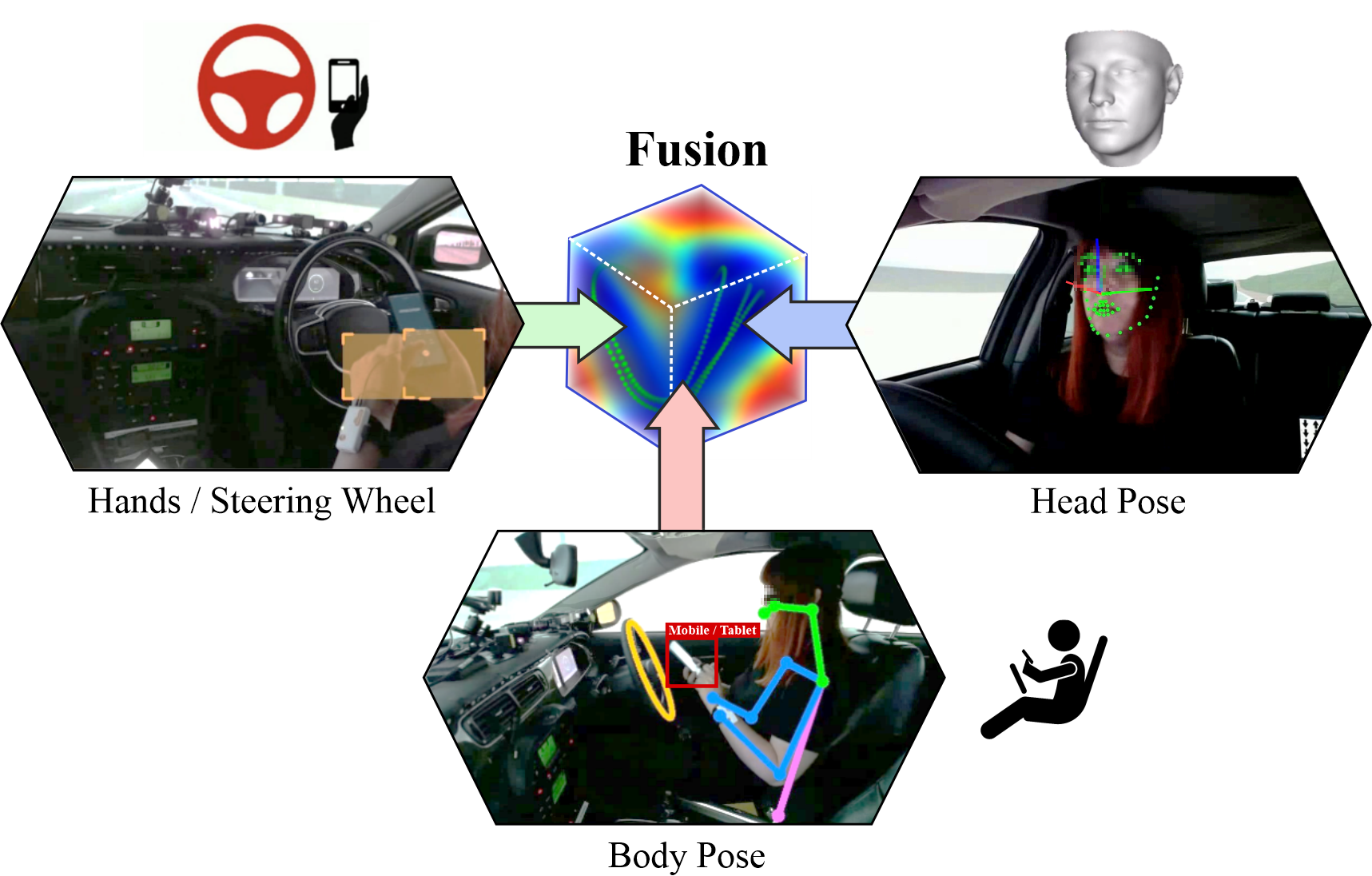}
   \vspace{-6mm}
   \caption{Proposed Cross-Modal fusion strategy for assessing driver situational awareness and readiness in response to critical take-over requests for \mbox{Level-3} automated vehicles.}
   \label{fusion}
   \vspace{-6mm}
\end{figure}
State-of-the-art research in DMS incorporates the latest developments in sensor technologies and utilises computer vision and machine learning techniques to achieve a robust and accurate driver state estimation. 
While advancements in autonomous vehicle technology have led to the deployment of robotaxis such as Waymo, in geofenced roads like San Francisco and Los Angeles, the primary focus of car manufacturers and academic researchers remains on enhancing the reliability of SAE Level 2 (L2) and Level 3 (L3) vehicles on public roads. According to the SAE automation levels \cite{SAE}, drivers in L2 vehicles with partial automation must continuously monitor the driving environment and retain full responsibility for vehicle control. In contrast, L3 vehicles, with conditional automation, allow drivers to disengage from active driving and engage in non-driving-related tasks (NDRTs) \cite{NDRT1}. However, this modal shift can diminish the driver’s vigilance, lower their situational awareness, and pose potential challenges during critical take-over requests (TOR) \cite{NDRT2, SA}. To address such challenges, DMS is tasked with continuously assessing the driver’s state in L3 vehicles to ensure safe and timely transitions from automated to manual control. This transition must safely occur within a few seconds, without worsening the complexity of the situation. If the DMS detects that the driver is not adequately prepared for the take-over, the system will initiate a minimum risk manoeuvre \cite{MRM}, to stop the car in a safe location.

Among various driver state monitoring solutions, computer vision-based models are proven to be a promising solution in Level 3 vehicles to estimate driver readiness in response to take-over requests \cite{Melcher2015}, particularly when the AV reaches the end of its operational design domain (ODD) or fails to operate in a complex scenario such as a road maintenance or lane blockage due to an accident. In the next section, we provide a literature review of the latest research in the field using in-cabin cameras and vision-based models.

\section{Literature Review}

Modern DMS consider hybrid solutions, by using facial feature analysis \cite{facial}, eye-tracking \cite{eye-gaze}, and head pose estimation \cite{head-pose}, to extract signs of driver distraction or fatigue. Convolutional Neural Networks (CNNs) \cite{CNN} and Recurrent Neural Networks (RNNs) \cite{RNN} have been extensively used for real-time video analysis to monitor yawning frequency or eyelid closures as key indicators of drowsiness. For example, Qu et al. developed a CNN-based spatio-temporal model to classify drivers' level of distraction, for a 10-class non-driving related task \cite{10class}. Similarly, Deo and Trivedi \cite{Deo2020}, along with Kazemi et al. \cite{kazemi2025}, used Long Short-Term Memory (LSTM) networks to estimate driver take-over readiness, by utilising in-cabin visual cues and expert-labelled Observable Readiness Index (ORI).

To overcome the limitations of single-modality input, multimodal frameworks have been proposed. These systems extract various data types from wearable sensors, in-cabin cameras, and vehicle dynamics to provide a comprehensive perspective of the driver’s state \cite{wearable}. 
For example, Abbas et al. \cite{multi-inattention} combined multi-view cameras and physiological sensors within a CNN-RNN hybrid to detect driver inattention. Deng et al. \cite{Deng2024} explored physiological responses such as heart rate (HR), skin conductance level (SCL), and mental workload (MWL), under simulated take-over conditions to quantify readiness delays. The "DeepTake" framework ~\cite{deeptake} extended this approach by predicting response time, intention, and take-over quality using a comprehensive set of vehicle and biometric features, although such wearable sensor-based systems face real-world scalability issues.

Some recent DMS research works have focused on near-infrared imaging and high-resolution cameras to detect micro-expressions and improve their robustness to lighting variations \cite{NIR}. Deep learning models with attention mechanisms \cite{attention} and explainable architectures \cite{explainable} are also being developed to improve the accuracy and interpretability of DMS. Furthermore, federated learning \cite{federated1}, \cite{federated2} has emerged to address in-cabin privacy concerns while maintaining high model performance across diverse user datasets.

\begin{figure}[t!]
\vspace{1mm}
   \centering
   \includegraphics[width = 0.85\linewidth]{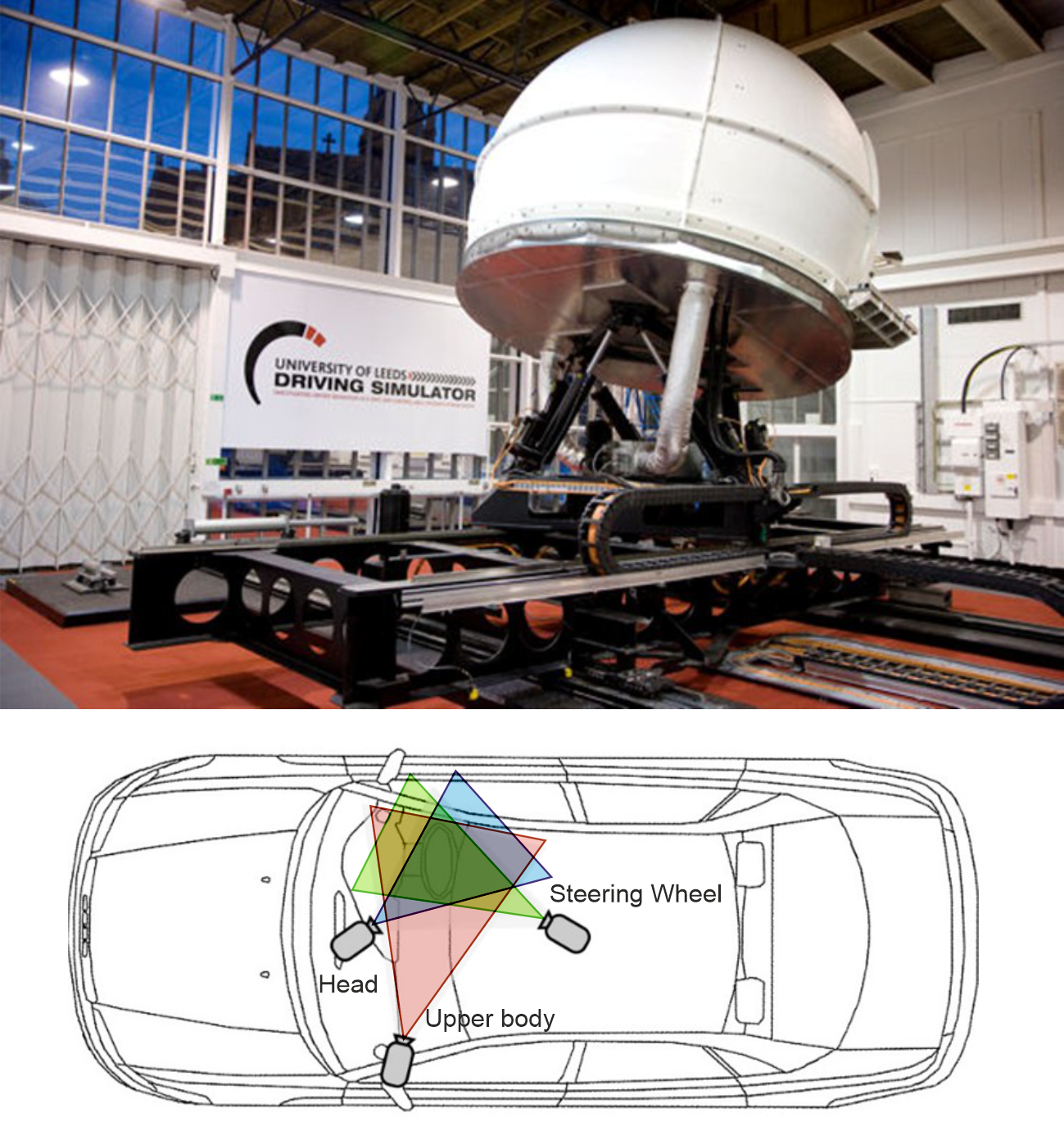}
   \vspace{-2mm}
   \caption{The University of Leeds Driving Simulator (top) with a \mbox{Triple-camera} setup for driver state monitoring (bottom) in a real Jaguar car inside the simulator dome.}
   \label{simulator}
   \vspace{-3mm}
\end{figure}
Several studies have focused on facial features and the driver's eye gaze. A recent study presented techniques for simultaneously detecting fatigue and distracted driving by utilising facial alignment networks to identify facial feature points and calculate distances to detect eye and mouth states, achieving improved accuracy and computation time compared to previous approaches \cite{inattention}. Bonyani et al. \cite{BONYANI} propose the DIPNet model to predict driver intention that can be used as a sign of the driver's state in response to a TOR. Braunagel \cite{Braunagel2018} explores an in-depth analysis of driver readiness in critical TOR via eye-on-the-road and gaze shift tracking in a simulator environment.
Lui et. al. \cite{Liu2024} propose the ACTNet model to predict minimum anticipated collision time (min ACT), as an indicator of drivers' readiness and performance for a take-over. The proposed CNN model mainly relies on eye-tracker data with a focus on eye fixation, blink rate, and pupil diameters as the main input features to determine driver readiness. However, reliance on eye-tracking data alone may not fully capture the multidimensional nature of readiness.

Other studies introduced large-scale datasets like the Driver Monitoring Dataset (DMD) \cite{dataset}, designed for benchmarking attention and drowsiness under various driving contexts. Transformer-based architectures \cite{transformers}, \cite{FP1, FP2} have also been applied to improve temporal feature modelling and cross-modal alignment.

%In a study by Ortega et al., a Driver Monitoring Dataset (DMD) was introduced for benchmarking driver drowsiness by encompassing real and simulated driving scenarios \cite{dataset}. Transformer-based architectures have also been applied to improve sequence modelling \cite{transformers} and feature representation in driver monitoring tasks \cite{FP1, FP2} to improve temporal feature modelling and cross-modal alignment. 
In another study by Kim et al. \cite{kim2022}, they propose a fundamental model for driver state estimation and investigate the influence of NDRTs on subjective driver readiness and take-over performance in L3 vehicles.

Despite these advancements, the reviewed works suffer from at least one of the following key limitations:

\begin{itemize}
    \item Mono-modality: Many systems rely exclusively on facial or head-based inputs, and ignore critical cues from body posture and hand activity, which are vital for assessing true readiness.
\end{itemize}

\begin{itemize}
    \item Sensor Practicality: High dependency on intrusive biosensors limits deployment feasibility in the AVs.
\end{itemize}

\begin{itemize}
    \item \noindent Spatio-Temporal Oversight: Few studies model readiness as a gradual, temporally evolving process. Our studies show that as the driver's attention reorients from non-driving tasks, the transition period typically spans over 1.6 seconds, to encompass steering wheel adjustments, situational scanning, and posture correction, which indicates the necessity of temporal modelling.
\end{itemize}

\noindent To address these gaps, we provide four key contributions as follows:

1- \textit{Multi-Camera Cross-Modal Fusion Architecture:} We propose a first-of-its-kind triple-camera driver monitoring system that combines front, side, and over-shoulder views. The fusion of head pose, hand position, and body posture significantly outperforms single-modality approaches in assessing readiness.

2- \textit{Spatio-Temporal Feature Integration:} Our method captures readiness as a dynamic, time-dependent process using synchronised video inputs and attention-based temporal modelling, moving beyond static indicators like gaze input.

3- \textit{High-Performance, Real-Time Readiness Classification:} Driver-Net achieves a very high accuracy, demonstrating strong generalisation across diverse driver profiles and activities. The system is also lightweight enough for real-time deployment in modern vehicles.

4- \textit{Safety-Critical Advancement:} By improving the accuracy and timing of take-over readiness detection, the proposed system supports safer transitions from automated to manual control, aligning with Euro NCAP and EU GSR 2024–2026 safety mandates and reducing risks associated with delayed driver responses.\\

\section {Methodology}
\begin{figure}
    \centerline{\includegraphics[width=1.02\linewidth]{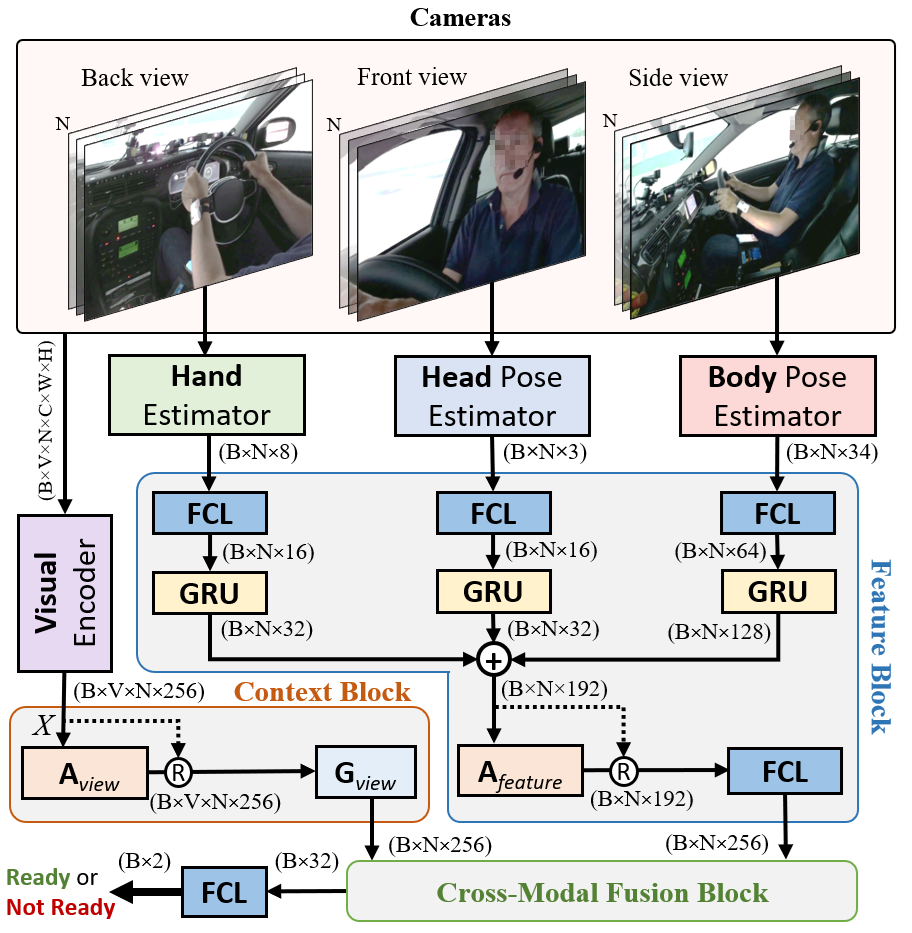}}
    \caption{The block diagram of the Driver-Net architecture, including camera input module and triple feature extractors, Visual encoder,  Feature Block, Context Block, and Cross-Modal Fusion block.
    \label{fig:model}}
    \vspace{-3mm}
\end{figure}

The proposed Driver-Net deep neural network architecture processes $N$-frame temporal video sequences captured at 30 FPS from three synchronised HD cameras ($1920 \times 1080$ pixels): front, side, and over-shoulder as depicted in Figure \ref{fig:model}. The Raw video frames are then converted to $720 \times 480$ pixels and are processed using three pre-trained feature extraction algorithms for hand detection, head-pose estimation, and body joint/pose estimation before being fed into the second layer of the model. The spatio-temporal video contents and extracted features are processed by two parallel `Context' and `Feature' blocks, and their outputs are fused by the `Cross-Modal Fusion' block to finally predict the driver's readiness for a take-over request. Figure \ref{fig:model} illustrates the model inputs, feature extractors, and the network architecture, including the input-output sizes of each module. 

\subsection{Context Block}
The Context Block is responsible for processing multi-view spatio-temporal features from input video sequences and aggregating them into a unified representation. A pre-trained ResNet-18 3D \cite{feichtenhofer2019slowfast} prepares the input of the Context Block, which serves as a visual encoder to extract spatio-temporal features from all three input videos.  The input tensor of the visual encoder has the shape $ \mathbb{R}^{B \times V \times N \times C \times W \times H}$, where $B$ is the batch size, $V$ is the number of camera views (front, side, over shoulder), $N$ is the temporal sequence length, $C$ is the number of colour channels (RGB), and $W$, $H$ are the spatial dimensions.
Since ResNet-18 3D requires input tensors of shape $ \mathbb{R}^{B \times N \times C \times W \times H}$, the camera views are flattened into the batch size dimension, resulting in input tensors with shape $ \mathbb{R}^{(B \cdot V) \times N\times C\times W \times H}$. After processing by the encoder, the output feature maps have shape $\mathbb{R}^{(B \cdot V) \times N \times 256}$, following global average pooling across the spatial dimensions $(W \times H)$. These feature maps are then reshaped to $\mathbb{R}^{B \times V \times N \times 256}$ for the subsequent layers.

A multi-head self-attention layer $\mathbf{A}_{\text{view}}$ operates on the dimension $V$, to capture inter-camera dependencies at each time step $N$ while preserving the spatial and temporal structure. 
The self-attention layer with $h=4$ heads can be formulated as:
\begin{equation}
\mathbf{A}_{\text{view}}(X) = (\text{Att}_1 \oplus \text{Att}_2 \oplus \dots \oplus \text{Att}_h)W_O \label{eq:att}
\end{equation}
\noindent where $\oplus$ denotes the concatenation operator, The input $X$ which is processed by the visual encoder with shape $\mathbb{R}^{B \times V \times N \times 256}$, and \( W_O \in \mathbb{R}^{(h \cdot d_v) \times 256} \) is a learnable weight matrix that projects the concatenated attention heads back to the original feature dimension of \( 256 \). Here, $d_v = 256 / h$, and the attention mechanism for $h$-th head, $\text{Att}_h$, is defined as follows:
\begin{equation}
\text{Att}_h(Q_h, K_h, V_h) = \text{softmax}\left(\frac{Q_h K_h^T}{\sqrt{d_k}}\right) V_h, 
\end{equation}
\noindent where the \textit{query}, \textit{key}, and \textit{value} projections for the $h$-th head are given by: 
\begin{equation}
 Q_h, K_h, V_h  = XW_h^Q, XW_h^K, XW_h^V
\end{equation}
\noindent \(W_h^Q\in \mathbb{R}^{256 \times d_k }, W_h^K\in \mathbb{R}^{256 \times d_k }, W_h^V\in  \mathbb{R}^{256 \times d_v }\) are the learnable projection matrices for queries, keys, and values, respectively.  
The input $X$ is reshaped from $\mathbb{R}^{B \times V \times N \times 256}$ into $\mathbb{R}^{(B \cdot N) \times V \times 256}$ to compute the attention over the $V$ dimension.
We also set  $d_k=d_v = 256/h$ to ensure consistency when concatenating the outputs of all heads.

To stabilise training and preserve gradient flow, a residual connection $\text{`R'}$ adds the input to the output of the attention layer (see dashed line arrow), followed by a layer normalisation \cite{ba2016layer}. 
The result is then passed to an aggregation module $\mathbf{G}_{\text{view}}$ that summarises the multi-view camera information across the dimension $V$, producing outputs of shape $\mathbb{R}^{B \times N\times 256}$. 
The following three common aggregation methods were evaluated to find the most suitable for the task:
\vspace{-2mm}
\begin{enumerate}
    \item \textbf{Global Average Pooling (GAP):} Computes the average features across the entire dimension $V$ at each time step in the input sequence with the length $N$, reducing feature complexity while preserving temporal dependencies.
    \vspace{-2mm}
    \begin{equation}
    \mathbf{G}_{\text{view}}^{\text{GAP}}(X_{\text{nr}}) = \frac{1}{V} \sum_{v=1}^{V} X_{\text{nr}, v}
    \end{equation}
    \noindent where $X_{\text{nr}}\in \mathbb{R}^{B \cdot N\times 256}$ is the normalised input to the aggregation module after the residual connection and layer normalisation.\\

    \item \textbf{Weighted Sum:} Applies a set of learnable weights $w_v$ to each camera view's features and sums them up.
    \vspace{-2mm}
    \begin{equation}
    \mathbf{G}_{\text{view}}^{\text{WS}}(X_{\text{nr}}) = \sum_{v=1}^{V} w_v \cdot X_{\text{nr}, v}
    \end{equation}
    \noindent where $w_v \in  \mathbb{R}^{V}$ are learned parameters which are normalised to satisfy $\Sigma^{V}_{v=1} w_v = 1$, ensuring interpretability and stability. \\

    \item \textbf{1D Convolution:} A 1D convolution is applied across the dimension $V$, producing a summarised output over the $V$ camera views.
    %
    %\vspace{-1mm}
    \begin{equation}
    \mathbf{G}_{\text{view}}^{\text{Conv1D}}(X_{\text{nr}}) = \text{Conv1D}(X_{\text{nr}})
    \end{equation}
    \noindent where $\text{Conv1D}$ uses convolutional filters of shape $(k,256)$, with the kernel size $k$ with stride one.
    % and stride parameters chosen based on the desired level of summarisation. 
    This method allows for learning spatial patterns across multiple camera inputs. %views. 
\\
\end{enumerate}
The Context Block's final output is a tensor $X_{\text{context}} \in \mathbb{R}^{B \times N \times 256}$ containing the processed spatio-temporal feature maps of the driver from three different camera views. The $X_{\text{context}}$ is one of two inputs of the Cross-Modal Fusion block.

\subsection{Feature Block}
\begin{figure*}[t!]
   \centering
   \vspace{2mm}
   %\vspace{-5mm}
   \includegraphics[width = 0.9\linewidth]{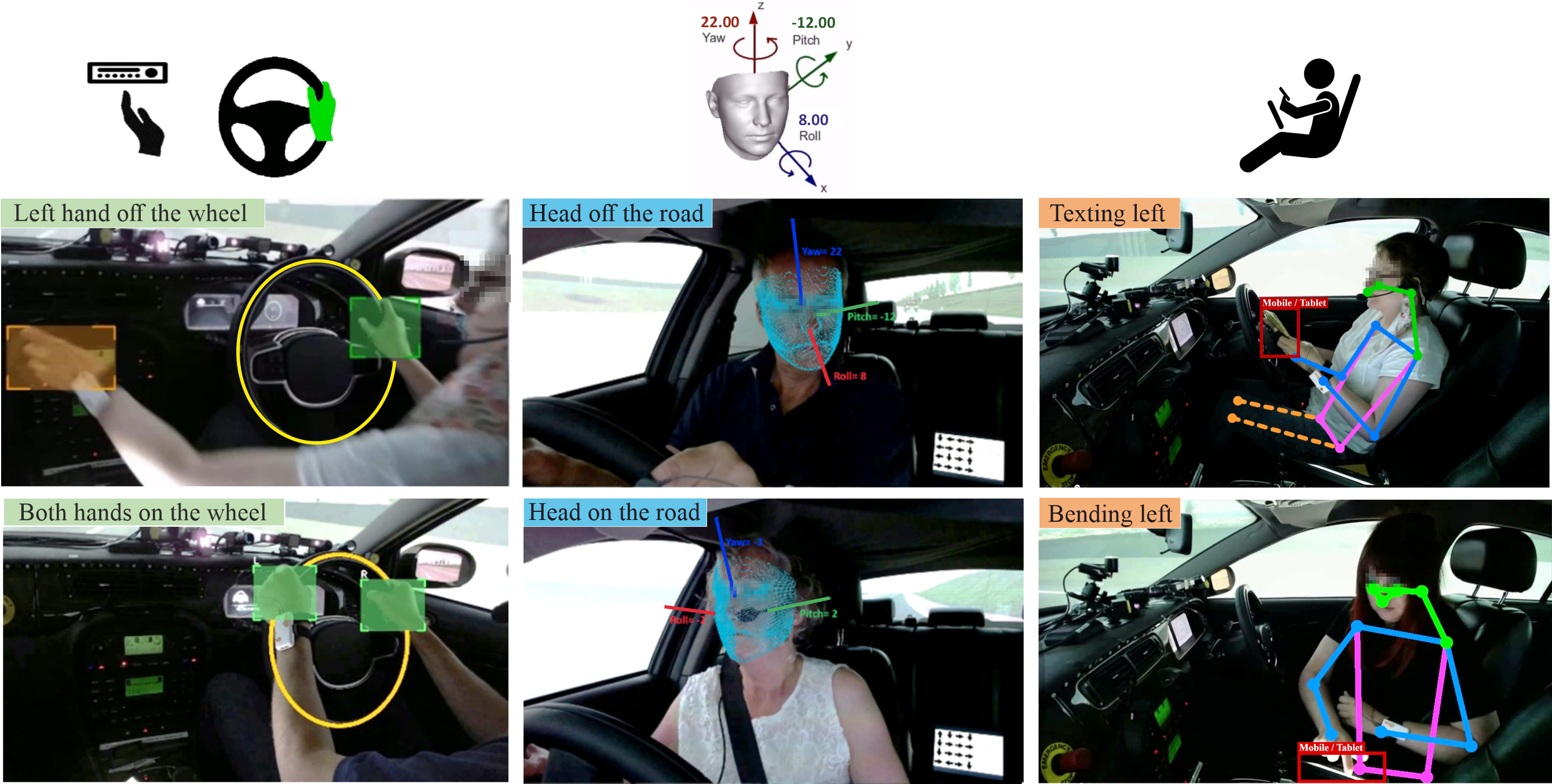}
   \vspace{-2mm}
   \caption{\textbf{Left}: Hands/steering wheel positioning and activity recognition. \textbf{Middle}: Head-pose estimation (Yaw, Pitch, Roll $ \rightarrow$ head on/off the road).\\ \textbf{Right}: Body pose estimation: normal, leaning/rotating, texting, and object detection.}
   \label{3-camera}
   \vspace{-3mm}
\end{figure*}

The Feature Block processes the input videos and extracts driver-centric features (see Figure \ref{3-camera}) using three computer vision algorithms as follows:\\
\begin{itemize}

    \vspace{-2mm}
    \item \textbf{Head Pose Estimator:} The 3-D direction of driver head attention, including Yaw, Pitch, and Roll angles, is estimated using 3DDFA \cite{wang20243d} from the frontal view camera. The collected features will be stored in the angles' tensor $X_{hd}\in \mathbb{R}^{B \times  N \times 3}$, to be used later in the Feature Block.\\
    \vspace{-2mm}
    
    \item \textbf{Body Pose Estimator:} Body joint features and their coordinates tensor $X_{bp}\in \mathbb{R}^{B \times  N \times 34}$ are extracted using YOLO7-Pose \cite{wang2023yolov7} from the side-view camera. Additionally, a YOLOv7 module detects objects within the region of the driver's hand, represented by the tensor $O \in \mathbb{R}^{B \times N \times M \times 2 \times 2}$, where $M$ is the number of detected objects per frame, and $2 \times 2$ 
    represents the dimension of the bounding box coordinates: top left corner $(x_t,y_t)$ and bottom right corner $(x_b, y_b)$.    \\
    \vspace{-2mm}
        
    \item \textbf{Hand Detector:} A YOLO7-based hand detection model \cite{wang2023yolov7} re-trained on the \textit{EgoHands} dataset \cite{Bambach_2015_ICCV}, and used to identify the position of the driver's hands relative to the steering wheel from the rear-view (over shoulder) camera. The hand detector generates a tensor of bounding box coordinates for the left $X_{hl}\in\mathbb{R}^{B \times  N \times 4}$ and right $X_{hr}\in\mathbb{R}^{B \times  N \times 4}$ hand. These will be then concatenated to $X_{hp}\in\mathbb{R}^{B \times  N \times 8}$
   
\end{itemize}
A fully connected layer (FCL) independently transforms these tensors into higher-dimensional spaces while preserving temporal structure. For the input feature tensor $X_f \in  \mathbb{R}^{(B \cdot  N) \times d_{in}}$,  which is flattened to the batch size ($B$) and the sequence length ($N$) dimension, the FCL uses a learnable weight matrix $W \in  \mathbb{R}^{(B \cdot  N) \times d_{out}}$ and a vector $b \in  \mathbb{R}^{d_{out}}$, producing the transformed output: \vspace{-2mm}

\begin{equation}
\text{FCL}(Y) = \text{ReLU}(X_fW  + b)
\vspace{-1mm}
\end{equation}

\noindent where $Y \in  \mathbb{R}^{(B \cdot  N) \times d_{out}}$ is the higher-dimensional representation of the input features, \text{ReLU} is a non-linear activation function \cite{nair2010rectified}, and $d_{\text{in}}, d_{\text{out}}$ are the input and output feature dimensions, respectively.
The $d_{in} \rightarrow d_{out}$ transformation is $34 \rightarrow 64$, $3 \rightarrow 16$, and $8 \rightarrow 16$, for the body pose, head pose; and  hand status, respectively. (See Figure~\ref{fig:model}).

Once FCL units process the body, head, and hand feature vectors, a Gated Recurrent Unit (GRU) \cite{chung2014empirical} captures temporal dependencies across the sequence of each tensor. 
Then, a concatenation operator combines the multiple GRU-processed tensors along with feature dimension, creating a new tensor of $\mathbb{R}^{B \times N \times 192}$. 

A feature-wise self-attention layer $\mathbf{A}_{\text{feature}}$ then operates on the feature dimension $(\text{i.e.}\, 192)$, and dynamically identifies the most relevant features while de-prioritising redundant ones. Residual connections and layer normalisation ensure stable gradient propagation. An additional FCL processes the output tensor to align feature dimensions for %downstream 
next modules.

The Feature Block's final output is a tensor $X_{\text{feature}} \in \mathbb{R}^{B \times N \times 256}$ containing the processed temporal feature maps of the driver's body pose, head angles, and hand position.

\subsection{Cross-Modal Fusion Block}
The Cross-Modal Fusion block integrates all spatio-temporal features from the Context and Feature blocks. Three widely used fusion methods in neural networks were evaluated:
\begin{enumerate}
    \item \textbf{Concatenation Fusion (CF):} $X_{\text{context}}$ and $X_{\text{feature}}$ are concatenated along the feature dimension, yielding $\mathbb{R}^{B \times N \times 512}$. An FCL is then applied to process and reduce the input's feature dimensionality from $\mathbb{R}^{B \times (N \cdot 512)}$ to  $\mathbb{R}^{B \times 256}$. The FCL processes both the temporal and feature dimensions along the batches. CF is widely used in multimodal fusion for its simplicity and ability to preserve modality-specific features \cite{zhao2024deep}.\\

    \vspace{-2mm}
    \item \textbf{Additive Fusion (AF):} This technique applies an element-wise summation of $X_{\text{context}}$ and $X_{\text{feature}}$ across the feature dimension, which results in a tensor $\mathbb{R}^{B \times N \times 256}$, followed by global average pooling across the temporal dimension $N$, producing a tensor with the shape $\mathbb{R}^{B \times 256}$. AF emphasises shared information and balances contributions of both modalities \cite{bayoudh2022survey}.\\

    \vspace{-2mm}
    \item \textbf{Cross-Attention Fusion (CAF):} A cross-attention module, as defined in Eq. \eqref{eq:att}, computes the weights for both Feature and Context blocks, with $X_{\text{context}}$ as the query and $X_{\text{feature}}$ as the key and value. This enables the model to focus on the most relevant features and maximise the accuracy \cite{zhang2023cross}. The outcome tensor is a fused representation of shape $\mathbb{R}^{B \times 256}$. CAF is particularly known for its effectiveness in fine-grained interaction modelling and has been widely applied in recent multimodal systems \cite{akbari2021vatt}.
\end{enumerate}

%\vspace{-2mm}
The output tensor of the fusion block is then processed and reduced to $\mathbb{R}^{B \times 32}$ via an FCL with a 50\% dropout rate to prevent overfitting \cite{srivastava2014dropout}. 

Ultimately, the last FCL is used for the final classification to predict binary labels (``ready" or ``not-ready") for the driver state in response to a TOR request.

\section{Experiments}
This section details the test environment, dataset, training and testing procedures, and the outcomes.
\noindent \subsection{\textbf{Test Environment:}}
We conducted a series of experiments to assess the impact of each module variation on the overall performance of the proposed model. These experiments involved training and testing various components — namely, the Context Block, Feature Block, and Cross-Modal Fusion block — using the University of Leeds Driving Simulator (UoLDS), currently the most advanced driving simulator in the UK. The simulator dome includes a real Jaguar car and offers almost equal degrees of movement freedom with an immersive feeling of driving in the real world. The Jaguar cabin is equipped with a triple-camera module, positioned to capture comprehensive in-cabin driver behaviour, as illustrated in Figure \ref{3-camera}:\\
\begin{itemize}
{
\vspace{-3mm}
    \item Frontal Camera: Positioned flexibly around the central dashboard, this camera faces the driver's head and feeds into the ``Head Pose Estimator", which infers the driver’s visual attention using yaw, pitch, and roll angles, as shown in Fig. \ref{3-camera}. It also tracks the duration of attention diversion, for example, by detecting sequences of $N$ consecutive frames where the head orientation is off-road.\\

\vspace{-3mm}
    \item Over-Shoulder Camera: Mounted to capture a rear view of the driver, this camera supports the ``Hand Detector" module in identifying the position and activity of both hands. It determines whether one or both hands are on the steering wheel, and if not, identifies their alternative locations and duration of disengagement—e.g., a hand on the gear lever or infotainment knob for $t$ seconds before returning to the wheel.  \\
\vspace{-3mm}
   
    \item Side Camera (Passenger-Side A-Pillar): This camera captures the driver’s full-body posture and can identify contextual cues, such as interactions with handheld devices. The Body Pose Estimator uses this input to assess seating alignment (e.g., upright vs. leaning or reaching postures). Additionally, this side view provides redundancy in hand localisation, particularly valuable when occlusion affects the over-shoulder view, such as with tall drivers or those with larger body mass.
}
\end{itemize}

\noindent \subsection{\textbf{Dataset:}}

We recruited 30 licensed drivers of various ages and sexes, all holding valid UK driving licences. The experiment was conducted in the University of Leeds Driving Simulator and encompassed both automated and manual driving modes on simulated multi-lane urban roads. During automated driving, participants were permitted to engage in NDRTs, such as using a tablet. In specific scenarios, participants received phone calls from the simulator control room and engaged in conversation while manually driving. Additionally, activities such as texting or browsing on a smartphone were intentionally introduced to simulate common distractions.

The UoLDS dataset consists of a selection of 540 video clips from each of the three in-cabin cameras, totalling 1620 videos. Clip durations range from 10 to 25 seconds, reflecting various driver behaviours across both automated and manual driving modes. The driver's readiness status may change multiple times over time in the selected video cuts. However, only the last frame of each video clip was subjectively annotated by two expert reviewers as either ``ready" or ``not ready" in relation to a take-over request (TOR) initiated approximately two seconds earlier. These annotations were suggested by a thorough review of the synchronised triple-camera recordings, taking into account the driver’s overall in-cabin behaviour and situational context in the critical seconds preceding the TOR. 

From the total of 1,620 annotated video samples, 654 instances (40.37\%) were labelled as \textit{ready}, and 966 instances (59.63\%) were marked as \textit{not ready}. These expert labels were subsequently used as ground truth for evaluating the model's key performance indicators (KPIs). We used 1222 videos for training and 398 videos for testing. For the training phase, 50.16\% of selected samples were labelled as ready, and 49.84\% as not ready, demonstrating a balanced distribution of each label. \\

\noindent \subsection{\textbf{Training and Test:}}

The model was trained and evaluated on a CUDA-enabled parallel computing platform equipped with an NVIDIA RTX A6000 GPU, 64GB of RAM, and an Intel Core i9-13900K 24-core processor.

The \textit{Context} and \textit{Feature} blocks were trained independently, each with a randomly initialised fully connected classification layer (FCL) to learn distinct representations. For the \textit{Cross-Modal Fusion} block, we removed the classification heads from the pre-trained \textit{Context} and \textit{Feature} blocks, froze their learned parameters, and then only trained the \textit{Cross-Modal Fusion} block with a randomly initialised classification head. Further implementation details are provided below.

\noindent \subsubsection{\textbf{Context Block Training}}
\begin{table}[h]
    \centering
    %\vspace{2mm}
    \caption{Performance of different aggregation modules ($\mathbf{G}_{\text{view}}$) for the `Context block' against accuracy (A), precision (P), and recall-rate (R) metrics.}
    \vspace{-2mm}
    \label{table:context}
    \begin{tabular}{lccccc}
    \noalign{\hrule height 1pt}
    \textbf{\#} & \textbf{Parameters} & \textbf{Aggregation} & \textbf{A(\%)} & \textbf{P(\%)} & \textbf{R(\%)} \\
    \noalign{\hrule height 1pt}
    $\alpha_{GAP}$ & 270,338 & GAP & 85.1 & 84.7 & 85.0 \\
    $\alpha_{WS}$ & 270,341 & WS & 84.5 & 84.2 & 84.4 \\
    $\alpha_{C1D}$ & 466,946 & Conv1D & \textcolor{blue}{\textbf{88.0}} & \textcolor{blue}{\textbf{87.4}} & \textcolor{blue}{\textbf{87.7}} \\
    \noalign{\hrule height 1pt}
    \end{tabular}
\end{table}
The Context Block is aimed at learning a compact and discriminative representation of the driver's behaviour from the last 1.6-second multi-view spatio-temporal features. The $X_{\text{context}}$ tensor was flattened over the temporal dimension %(N=16, 10fps) 
and connected directly to the classification head (final Fully Connected Layer, FCL) to predict driver readiness. Our experiments show that with a temporal dimension of $N=16$ (10 \textit{fps} sampling), we can gain pretty much the same accuracy as the full-frame video input (30 \textit{fps}). 
We evaluated three aggregation modules-- Global Average Pooling (GAP), Weighted Sum (WS), and 1D Convolution (Conv1d) to aggregate the temporal features before classification. The Context block variants and their performance metrics on the test set are presented in Table \ref{table:context}. 
\begin{table}[h]
    \centering
    \vspace{-2mm}
    \caption{Performance of every single feature and all features combined, the `Feature Block' against accuracy (A), precision (P), and recall-rate (R) metrics.}
    \vspace{-2mm}
    \label{table:feature}
    \begin{tabular}{lccccc}
    \noalign{\hrule height 1pt}
    \textbf{\#} & \textbf{Parameters} & \textbf{Feature set} & \textbf{A(\%)} & \textbf{P(\%)} & \textbf{R(\%)} \\
    \noalign{\hrule height 1pt}
    $\beta_P$ & 32,634 & Body Pose & 71.1 & 70.56 & 70.8 \\
    $\beta_A$ & 21,370 & Head Angles & 43.2 & 42.68 & 43.1 \\
    $\beta_H$ & 19,450 & Hand Position & 69.3 & 69.01 & 68.7 \\
    \hdashline
    $\beta_{all}$ & 184,978 & All Features & \textcolor{blue}{\textbf{86.94}} & \textcolor{blue}{\textbf{86.30}} & \textcolor{blue}{\textbf{86.50}} \\
    \noalign{\hrule height 1pt}
    \end{tabular}
\end{table}
\noindent \subsubsection{\textbf{Feature Block Training}}

We trained the Feature Block aimed at the temporal learning of the driver's state by integrating head pose, body pose, and hand position features extracted from a triple-camera setup. Similar to the Context Block, and in order to reduce the computation cost, the $X_{\text{feature}}$ tensor was flattened over the dimension of $N=16$ and connected directly to the classification head to estimate the driver readiness state. We evaluated the performance of each feature set and the combined feature sets. The results are summarised in Table \ref{table:feature}, which proves the critical role of a triple-camera setup with a dramatic increase in the model performance in estimating driver readiness state.
\begin{table}[h]
\centering
    \caption{Performance of different fusion strategies for the Cross-Modal Fusion block.}
    \vspace{-2mm}
    \label{table:fusion}
    \begin{tabular}{lcccccc}
    \noalign{\hrule height 1pt}
    \textbf{Model} & \hspace{-2mm}\textbf{Parameters} & \hspace{-1mm}\textbf{Fusion} & \hspace{-1mm}\textbf{A(\%)} & \hspace{-1mm}\textbf{P(\%)} & \hspace{-1mm}\textbf{R(\%)} & \hspace{-1mm}\textbf{Cost} \\
    \noalign{\hrule height 1pt}
%
    %\multirow{3}{*}{$\alpha_{GAP}$, $\beta_{all}$} 
    %& 2,536,592 & CF & 88.4 & 88.1 & 88.3 & 19 \\
    %& 447,152 & AF & 90.8 & 91.8 & 90.7 & 44 \\
    %& 1,036,976 & CAF & 92.8 & 93.1 & 91.8 & 28 \\
    %\hline
 %   
    \multirow{3}{*}{$\alpha_{GAP}$, $\beta_{all}$} 
    & 2,536,592 & CF & \hspace{-1mm}88.4 & \hspace{-1mm}88.1 & \hspace{-1mm}88.3 & \hspace{-1mm}52ms\\
    & 447,152 & AF & \hspace{-1mm}90.8 & \hspace{-1mm}91.8 & \hspace{-1mm}90.7 & \hspace{-1mm}22ms \\
    & 1,036,976 & CAF & \hspace{-1mm}92.8 & \hspace{-1mm}93.1 & \hspace{-1mm}91.8 & \hspace{-1mm}35ms \\
    \hline
    
    \multirow{3}{*}{$\alpha_{WS}$, $\beta_{all}$} 
    & 2,536,595 & CF & \hspace{-1mm}87.9 & \hspace{-1mm}87.3 & \hspace{-1mm}87.4 & \hspace{-1mm}52ms \\
    & 447,155 & AF & \hspace{-1mm}89.1 & \hspace{-1mm}88.8 & \hspace{-1mm}89.0 & \hspace{-1mm}22ms \\
    & 1,036,979 & CAF & \hspace{-1mm}90.8 & \hspace{-1mm}91.8 & \hspace{-1mm}90.8 & \hspace{-1mm}35ms \\
    \hline
    \multirow{3}{*}{$\alpha_{C1D}$, $\beta_{all}$} 
    & 2,741,394 & CF & \hspace{-1mm}\textcolor{blue}{\textbf{95.1}} & \hspace{-1mm}\textcolor{blue}{\textbf{95.6}} &\hspace{-1mm}\textcolor{blue}{\textbf{95.1}} & \hspace{-1mm}62ms \\
    & 651,954 & AF & \hspace{-1mm}91.2 & \hspace{-1mm}91.9 & \hspace{-1mm}\textcolor{purple}{\textbf{92.0}} & \hspace{-1mm}24ms \\
    & 1,241,778 & CAF & \hspace{-1mm}\textcolor{purple}{\textbf{93.2}} & \hspace{-1mm}\textcolor{purple}{\textbf{93.7}} & \hspace{-1mm}91.5 & \hspace{-1mm}37ms \\
    \noalign{\hrule height 1pt}
    \end{tabular}
    \vspace{-4mm}
\end{table}
\noindent \subsubsection{\textbf{Cross-Modal Fusion Block}}

Finally, we trained and evaluated the \textit{Cross-Modal Fusion} block by combining the pre-trained \textit{Context} and \textit{Feature} Blocks. Three different fusion strategies were trained: Concatenation Fusion (CF), Additive Fusion (AF), and Cross-Attention Fusion (CAF). The results are presented in Table \ref{table:fusion}.

Table \ref{table:fusion} shows that Concatenation Fusion and Cross-Attention Fusion perform best and second best, respectively, with the 1D convolution (in the Context block) and utilisation of all features (from the Feature block). Furthermore, the computational cost column ranging from 22ms to 62ms shows that the proposed model can process the input videos, ranging from 16 to 45 \textit{fps}, which is fast enough for the intended real-world application of TOR.

To evaluate the generalisation capability of the proposed models, we performed 5-fold cross-validation ($k=5$) on the entire dataset (Table \ref{tab:crossval_results}). Two training strategies were examined: $\mathcal{T}_{\text{all}}$, where the entire model including Context Block ($\alpha_{C1D}$), Feature Block ($\beta_{all}$), and Fusion Block were trained end-to-end and from scratch; and $\mathcal{T}_{\text{fusion}}$, where the Context block and Feature Block were frozen after pre-training, and only the Fusion Block was trained. As shown in Table~\ref{tab:crossval_results}, $\mathcal{T}_{\text{all}}$ consistently outperforms $\mathcal{T}_{\text{fusion}}$ in terms of accuracy, precision, and recall in line with the results obtained in Table \ref{table:fusion}. Similarly, the Concatenation Fusion (CF) strategy performed best; however, Additive Fusion (AF) took the second-best place, despite a significantly smaller number of parameters. While these confirm the critical role of Fusion Block in the model's overall performance, it may indicate that AF's simplicity allows a more effective generalisation across multiple folds when the model is trained from scratch, while CF and CAF's complexity might lead to overfitting within individual folds of cross-validation.
\begin{table}[h!]
\centering
\caption{Performance evaluation using 5-fold cross-validation (k=5). %$\mathcal{T}_{\text{all}}$: Model training on all three blocks of the model; $\mathcal{T}_{\text{fusion}}$: Training the Fusion Block only with frozen Context $\alpha_{C1D}$ and Feature $\beta_{all}$ blocks.
%\vspace{-2mm}
}
\label{tab:crossval_results}
\begin{tabular}{@{}lcccc@{}}
\noalign{\hrule height 1pt}
\textbf{Train} & \textbf{Fusion} & \textbf{A $\pm  \sigma$ (\%)} & \textbf{P $\pm  \sigma$ (\%)} & \textbf{R $\pm  \sigma$ (\%)} \\
\noalign{\hrule height 1pt}
$\mathcal{T}_{\text{all}}$ & CF  & \textcolor{blue}{\textbf{95.4}} $\pm$ 0.3 & \textcolor{blue}{\textbf{95.8}} $\pm$ 0.4 & \textcolor{blue}{\textbf{95.3}} $\pm$ 0.3 \\
$\mathcal{T}_{\text{all}}$ & AF & \textcolor{purple}{\textbf{93.9}} $\pm$ 0.4  
&  \textcolor{purple}{\textbf{94.3}} $\pm$ 0.5 &  \textcolor{purple}{\textbf{93.2}} $\pm$ 0.4 \\
$\mathcal{T}_{\text{all}}$ & CAF & $93.5 \pm 0.5$ & $94.0 \pm 0.6$ & 91.8 $\pm$ 0.5 \\
\hline
$\mathcal{T}_{\text{fusion}}$ & CF  & $93.1 \pm 0.4$ & $93.6 \pm 0.5$ & $92.9 \pm 0.6 $\\
$\mathcal{T}_{\text{fusion}}$ & AF & $ 92.3 \pm 0.4$ 
& $ 92.7 \pm 0.5 $ & $ 91.6 \pm 0.4 $ \\
$\mathcal{T}_{\text{fusion}}$ & CAF & $91.7 \pm 0.6$ & $92.3 \pm 0.6$ & $91.0 \pm 0.5$ \\
\noalign{\hrule height 1pt}
\end{tabular}
\vspace{-4mm}
\end{table}
\section {Conclusion and Future Directions}
This study presented a novel framework for estimating driver readiness in response to take-over requests (TOR) by leveraging multi-view spatio-temporal features extracted from a triple-camera setup and a quasi-naturalistic dataset collected from the University of Leeds Driving Simulator (UoLDS). The proposed model integrated a Context Block, Feature Block, and Cross-Modal Fusion block to capture and fuse in-cabin behavioural cues, to enhance take-over readiness estimation.

Experimental results highlighted the effectiveness of each model component and the overall fused approach. The Context Block, which extracts multi-view contextual features, achieved the highest performance when using a 1D convolutional aggregation strategy, reaching an accuracy of 88.0\%. The Feature Block, which utilises body pose, head angles, and hand position features, demonstrated significant improvements when combining all three modalities, achieving an accuracy of 86.94\%. Finally, the Cross-Modal Fusion Block, responsible for merging contextual and feature representations, led to noticeable improvement and achieved the highest overall accuracy of 95.8\% using a concatenation-based fusion strategy with Conv1D-driven context extraction.
These findings suggest that the proposed approach offers a promising direction for driver monitoring systems, particularly in Level 3 conditionally automated vehicles, where accurate estimation of take-over readiness is critical for safety and performance. 

While Driver-Net demonstrated strong performance and generalisation in estimating driver readiness using a multi-camera fusion strategy, the system is evaluated primarily within a high-fidelity simulator environment, which may not fully capture the variability and unpredictability of real-world driving conditions. 
Future work can focus on validating the model in real-world driving scenarios with a broader demographic pool. Incorporating non-intrusive physiological indicators and investigating continuous readiness scoring rather than binary classification may also enhance practical deployment and improve robustness and scalability.

\section*{Declarations}
The authors declare no conflicts of interest and no financial or personal relationships that could potentially influence the interpretation or presentation of the findings. 

\section*{ACKNOWLEDGMENT}
This research has received funding from the European Union’s Horizon 2020 research and innovation programme, for the Hi-Drive project under grant Agreement No 101006664. The article reflects only the author’s view, and neither the European Commission nor CINEA is responsible for any use that may be made of the information this document contains.

%\break

\end{document}